\documentclass[10pt, twocolumn]{article}

\usepackage[margin=2cm]{geometry}
\usepackage{setspace}
\usepackage{parskip}

\usepackage{amsmath, amssymb}
\usepackage{wasysym}

\usepackage{tikz}
\usetikzlibrary{arrows.meta, positioning, shapes.geometric, calc, decorations.pathreplacing, backgrounds, fit}
\usepackage{pgfplots}
\pgfplotsset{compat=1.18}
\usepackage{graphicx}
\usepackage{float}
\usepackage{caption}
\usepackage{subcaption}
\usepackage{booktabs}
\usepackage{multirow}
\usepackage{array}

\usepackage{xcolor}
\definecolor{layerA}{RGB}{220, 50, 50}
\definecolor{layerB}{RGB}{50, 120, 200}
\definecolor{layerC}{RGB}{100, 160, 80}
\definecolor{primary}{RGB}{41, 65, 122}
\definecolor{accent}{RGB}{200, 80, 60}
\definecolor{neutral}{RGB}{120, 120, 120}

\usepackage[colorlinks=true, linkcolor=primary, citecolor=accent, urlcolor=primary]{hyperref}
\usepackage[numbers, sort&compress]{natbib}

\usepackage{enumitem}
\usepackage{algorithm}
\usepackage{algorithmic}
\usepackage{fancyhdr}
\usepackage{titlesec}
\titleformat{\section}{\Large\bfseries\color{primary}}{\thesection}{1em}{}
\titleformat{\subsection}{\large\bfseries\color{primary!80!black}}{\thesubsection}{1em}{}
\titleformat{\subsubsection}{\normalsize\bfseries\color{primary!60!black}}{\thesubsubsection}{1em}{}

\title{
    {\LARGE\bfseries\color{primary} AI-Driven Multi-Agent Simulation of Stratified Polyamory Systems:} \\[0.3cm]
    {\large A Computational Framework for Optimizing Social Reproductive Efficiency}
}
\author{
    \textsc{Yicai Xing} \\[0.2cm]
    {\small Independent Researcher} \\[0.1cm]
    {\small\texttt{xingyc18@tsinghua.org.cn}}
}
\date{}

\begin{document}

\twocolumn[
\maketitle
\thispagestyle{fancy}

\begin{abstract}
\noindent Contemporary societies face a severe crisis of demographic reproduction. Global fertility rates continue to decline precipitously, with East Asian nations exhibiting the most dramatic trends---China's total fertility rate (TFR) fell to approximately 1.0 in 2023, while South Korea's dropped below 0.72 \cite{un2024,choe2025}. Simultaneously, the institution of marriage is undergoing structural disintegration: educated women rationally reject unions lacking both emotional fulfillment and economic security, while a growing proportion of men at the lower end of the socioeconomic spectrum experience chronic sexual deprivation, anxiety, and learned helplessness \cite{ueda2020,abdellaoui2025}. Monogamy, as the institutional complement to private property \cite{engels1884}, is seeing its historical rationale erode. This paper proposes a \textbf{computational framework} for modeling and evaluating a \textbf{Stratified Polyamory System (SPS)} using techniques from \textbf{agent-based modeling (ABM)} \cite{epstein1996,bonabeau2002}, \textbf{multi-agent reinforcement learning (MARL)} \cite{lerer2017,zheng2022}, and \textbf{large language model (LLM)-empowered social simulation} \cite{park2023,gao2023}. The SPS permits individuals to maintain a limited number of legally recognized secondary partners (``companions'') in addition to one primary spouse, combined with socialized child-rearing and inheritance reform. We formalize the A/B/C stratification as heterogeneous agent types in a multi-agent system, where each agent possesses attributes (mate value, economic resources, attractiveness, fertility) and the SPS rules constitute the environment dynamics. The matching process is modeled as a \textbf{multi-agent reinforcement learning problem} amenable to Proximal Policy Optimization (PPO) and the mating network is analyzed using \textbf{graph neural network (GNN)} representations. Drawing on evolutionary psychology \cite{buss1993,trivers1972,buss2019}, behavioral ecology \cite{betzig1986,emlen1977}, social stratification theory \cite{becker1973,greenwood2014}, computational social science \cite{epstein1996,schelling1971}, algorithmic fairness \cite{mehrabi2021,roth2002}, and institutional economics, and invoking comparative anthropological evidence \cite{murdock1967,goldstein1987,gough1959} alongside contemporary consensual non-monogamy (CNM) research \cite{conley2013,moors2017,rubel2015,sheff2014}, we argue that SPS can improve aggregate social welfare in the Pareto sense. We present a simulation architecture and preliminary computational results demonstrating the framework's viability. The framework addresses the dual crisis of female motherhood penalties \cite{budig2001,england2016} and male sexlessness \cite{ueda2020,donnelly2001}, while offering a non-violent mechanism for wealth dispersion analogous to the historical Chinese ``Grace Decree'' (\textit{Tui'en Ling}) \cite{scheidel2017,piketty2014}.

\vspace{0.3cm}
\end{abstract}
\vspace{1cm}
]

\section{Introduction: The Structural Crisis of Monogamy and the Case for Computational Modeling}

\subsection{A Historical Materialist Examination of Marriage}

Engels, in \textit{The Origin of the Family, Private Property and the State}, argued that monogamy did not arise from natural sexual love but from economic conditions---specifically, from the triumph of private property over primitive group marriage \cite{engels1884}. Morgan's anthropological fieldwork among the Iroquois and other pre-state societies provided crucial evidence: group marriage and matrilineal inheritance coexisted without social dysfunction \cite{morgan1877}. The core function of marriage has never been the institutionalization of romantic love; rather, it has served as a \textbf{legal conduit for property transmission} and an \textbf{organizational unit for labor force reproduction}.

Henrich, Boyd, and Richerson (2012) offered a sophisticated complementary argument from the perspective of cultural evolution: normative monogamy prevailed in cross-cultural competition not because it is more ``natural,'' but because it suppresses intrasexual competition among males, thereby reducing intra-group violence and crime rates and conferring competitive advantages upon groups that adopted this institution \cite{henrich2012}. In other words, monogamy is a product of \textbf{cultural group selection}, not an individually optimal strategy. Scheidel (2009) further documented how Greco-Roman monogamy represented a ``peculiar institution'' in global context, reinforcing the view that strict pair-bonding norms are culturally contingent rather than biologically determined \cite{scheidel2009}.

However, contemporary transformations in productive forces have fundamentally undermined the compatibility between this institution and the prevailing relations of production:

\begin{enumerate}[leftmargin=1.5em]
    \item \textbf{Female economic independence} has weakened the necessity of marriage as an economic security mechanism. Oppenheimer's (1997) ``independence hypothesis'' posits that rising female income does not cause marriage decline per se, but it decisively alters mate selection criteria---women no longer ``settle'' due to economic dependence \cite{oppenheimer1997};
    \item \textbf{The knowledge economy} has dramatically escalated the costs of labor force reproduction. Folbre's (2001, 2008) economic analyses demonstrate that the full lifecycle cost of raising a middle-class child now far exceeds that of the industrial era \cite{folbre2001,folbre2008}, transforming children from assets into liabilities;
    \item \textbf{The Second Demographic Transition} (SDT): Van de Kaa (1987) and Lesthaeghe (2010, 2014) identified a fundamental values shift in post-industrial societies---from survivalism to self-expression---that has transformed attitudes toward fertility and marriage \cite{vandekaa1987,lesthaeghe2014};
    \item \textbf{Digital mating markets} have exposed the extreme inequality in the distribution of sexual attractiveness \cite{bruch2018,leichliter2010}, exacerbating structural imbalances in the marriage market.
\end{enumerate}

\begin{figure}[H]
\centering
\begin{tikzpicture}
\begin{axis}[
    width=\columnwidth, height=5.5cm,
    xlabel={Year},
    ylabel={Total Fertility Rate (TFR)},
    xmin=1960, xmax=2025,
    ymin=0, ymax=7,
    legend style={at={(0.98,0.98)}, anchor=north east, font=\tiny},
    grid=major,
    grid style={gray!20},
    tick label style={font=\tiny},
    label style={font=\small},
]
\addplot[thick, layerA, mark=*, mark size=1pt, mark repeat=5] coordinates {
    (1960,5.7) (1965,6.1) (1970,5.8) (1975,3.6) (1980,2.6) (1985,2.2)
    (1990,2.3) (1995,1.7) (2000,1.5) (2005,1.5) (2010,1.5) (2015,1.6)
    (2020,1.3) (2023,1.0)
};
\addlegendentry{China}
\addplot[thick, layerB, mark=square*, mark size=1pt, mark repeat=5] coordinates {
    (1960,6.0) (1965,5.0) (1970,4.5) (1975,3.4) (1980,2.8) (1985,1.7)
    (1990,1.6) (1995,1.6) (2000,1.5) (2005,1.1) (2010,1.2) (2015,1.2)
    (2020,0.84) (2023,0.72)
};
\addlegendentry{South Korea}
\addplot[thick, layerC, mark=triangle*, mark size=1pt, mark repeat=5] coordinates {
    (1960,2.0) (1965,2.1) (1970,2.1) (1975,1.9) (1980,1.8) (1985,1.8)
    (1990,1.5) (1995,1.4) (2000,1.4) (2005,1.3) (2010,1.4) (2015,1.5)
    (2020,1.3) (2023,1.2)
};
\addlegendentry{Japan}
\addplot[dashed, black, thick] coordinates {(1960,2.1) (2025,2.1)};
\node[font=\tiny, anchor=west] at (axis cs:2005,2.35) {Replacement: 2.1};
\end{axis}
\end{tikzpicture}
\caption{Long-term decline in total fertility rates (TFR) across major East Asian nations. Data: UN Population Division \cite{un2024}, national statistical bureaus, Choe \& Lee \cite{choe2025}.}
\label{fig:tfr}
\end{figure}
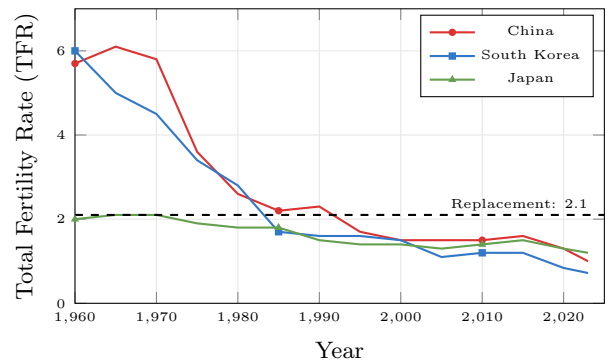

\subsection{The Dual Crisis of Contemporary Mating}

The contemporary mating predicament exhibits sharply gendered characteristics, though its root cause is unified---\textbf{monogamy produces severe resource misallocation when confronted with the highly unequal distribution of sexual attractiveness}.

\textbf{The female dimension}: Budig and England's (2001) seminal study demonstrated that each additional child reduces a mother's wages by approximately 7\% (the ``motherhood penalty'') \cite{budig2001}. England et al.\ (2016) further showed that this penalty persists even among highly skilled, highly paid women \cite{england2016}. Correll, Benard, and Paik (2007) experimentally confirmed systematic employer discrimination against mothers: with identical resumes, women with children were significantly less likely to be hired \cite{correll2007}. Under these institutional conditions, educated women's reluctance to marry men of comparable socioeconomic status is not a moral failing of ``excessive standards'' but \textbf{the inevitable outcome of rational cost-benefit analysis}.

\textbf{The male dimension}: A substantial proportion of men at the lower end of the socioeconomic hierarchy exist in a state of chronic sexual deprivation, social discipline, and low self-efficacy. Longitudinal data from the U.S.\ General Social Survey (GSS) reveal that the proportion of men aged 18--30 reporting no sexual activity in the past year rose from approximately 10\% in 2008 to approximately 28\% in 2018 \cite{ueda2020}. Abdellaoui et al.\ (2025) conducted a large-scale study linking sexlessness to adverse physical, cognitive, and personality traits \cite{abdellaoui2025}. Donnelly et al.\ (2001) provided a life-course analysis of involuntary celibacy, documenting its cascading psychological effects \cite{donnelly2001}. Moseley et al.\ (2025) proposed a dual pathways hypothesis connecting involuntary celibacy to both internalizing harm (depression, suicidality) and externalizing harm (radicalization, violence) \cite{moseley2025}. Case and Deaton's (2015, 2020) documentation of ``deaths of despair''---rising mortality among middle-aged white men due to substance abuse, alcohol, and suicide---is highly correlated with this social marginalization \cite{case2015,case2020}. Hudson and den Boer (2004) directly linked the ``bare branches'' crisis (surplus unmarried males) to social violence, crime rates, and political instability \cite{hudson2004}. Carter and Kushnick (2018) further confirmed the cross-cultural relationship between male aggressiveness and intrasexual contest competition \cite{carter2018}.

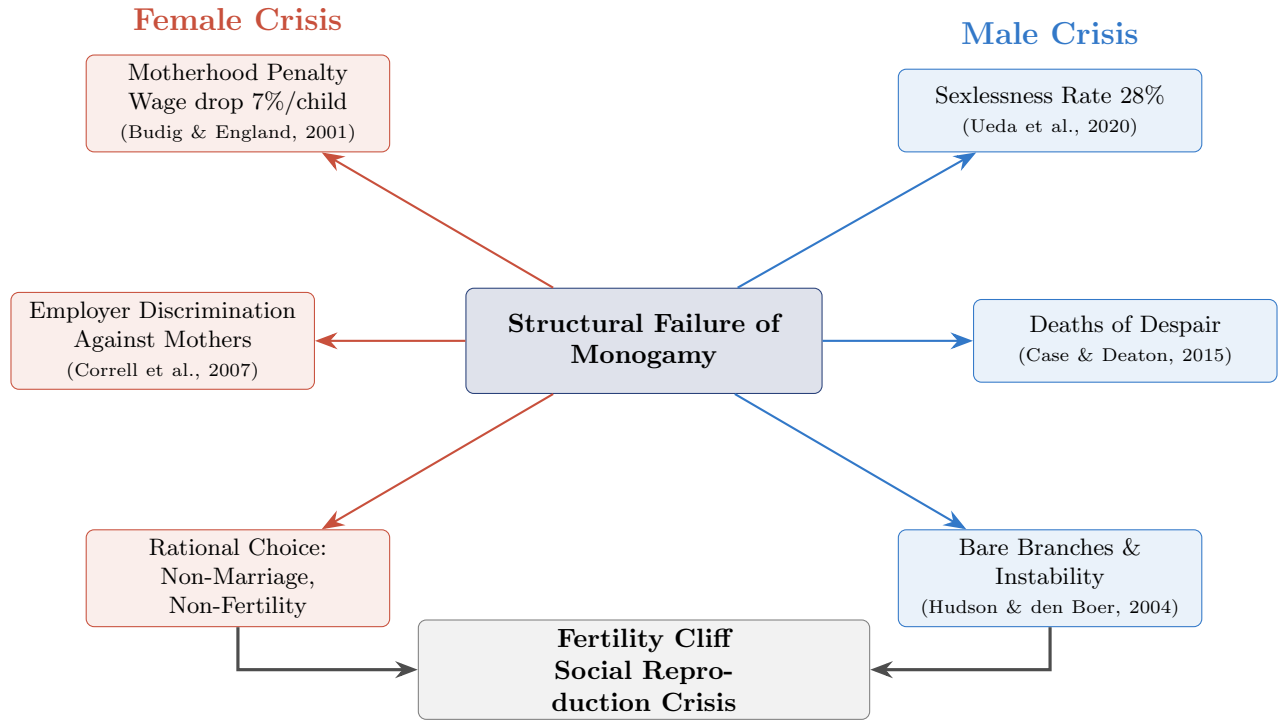
\begin{figure*}[t]
\centering
\begin{tikzpicture}[
    every node/.style={font=\small},
    box/.style={draw, rounded corners=3pt, minimum width=3.8cm, minimum height=1.1cm, align=center, text width=3.8cm},
    arrow/.style={-{Stealth[length=3mm]}, thick}
]
\node[box, fill=primary!15, draw=primary, text width=4.5cm, minimum height=1.4cm, font=\normalsize\bfseries] (center) {Structural Failure of\\Monogamy};

\node[box, fill=accent!10, draw=accent, above left=1.8cm and 1cm of center] (f1) {Motherhood Penalty\\Wage drop 7\%/child\\\scriptsize{(Budig \& England, 2001)}};
\node[box, fill=accent!10, draw=accent, left=2cm of center] (f2) {Employer Discrimination\\Against Mothers\\\scriptsize{(Correll et al., 2007)}};
\node[box, fill=accent!10, draw=accent, below left=1.8cm and 1cm of center] (f3) {Rational Choice:\\Non-Marriage, Non-Fertility};

\node[box, fill=layerB!10, draw=layerB, above right=1.8cm and 1cm of center] (m1) {Sexlessness Rate 28\%\\\scriptsize{(Ueda et al., 2020)}};
\node[box, fill=layerB!10, draw=layerB, right=2cm of center] (m2) {Deaths of Despair\\\scriptsize{(Case \& Deaton, 2015)}};
\node[box, fill=layerB!10, draw=layerB, below right=1.8cm and 1cm of center] (m3) {Bare Branches \&\\Instability\\\scriptsize{(Hudson \& den Boer, 2004)}};

\node[font=\large\bfseries, color=accent, above=0.2cm of f1] {Female Crisis};
\node[font=\large\bfseries, color=layerB, above=0.2cm of m1] {Male Crisis};

\draw[arrow, accent] (center) -- (f1);
\draw[arrow, accent] (center) -- (f2);
\draw[arrow, accent] (center) -- (f3);
\draw[arrow, layerB] (center) -- (m1);
\draw[arrow, layerB] (center) -- (m2);
\draw[arrow, layerB] (center) -- (m3);

\node[box, fill=gray!10, draw=gray, below=3cm of center, minimum width=6cm, font=\bfseries] (result) {Fertility Cliff\\Social Reproduction Crisis};
\draw[arrow, gray!60!black, very thick] (f3) |- (result);
\draw[arrow, gray!60!black, very thick] (m3) |- (result);
\end{tikzpicture}
\caption{The dual dimensions of monogamy's structural failure and their convergent effect on social reproduction. Each dimension is supported by key empirical literature.}
\label{fig:dual-crisis}
\end{figure*}

\subsection{Statement of the Problem}

Existing remedial measures---whether pro-natalist exhortation, fertility subsidies, or moral suasion---have uniformly failed to address the root cause. The Nordic countries' generous welfare-based fertility support policies, after producing initial gains, have seen fertility rates resume their decline \cite{lesthaeghe2014,gauthier2007}. Hoem (1993) documented the temporary nature of Sweden's policy-induced fertility bump \cite{hoem1993}. Hungary's aggressive pro-natalist incentives since 2019 (tax reductions, loan forgiveness) have produced only marginal short-term effects. Cowan and Wyndham-Douds (2022) found that even universal cash transfers have limited effects on fertility decisions \cite{cowan2022}. The central thesis of this paper is that a \textbf{constrained, stratified polyamory system} (Stratified Polyamory System, SPS), combined with socialized child-rearing and inheritance reform, can systematically alleviate the multiple crises confronting contemporary mating institutions. Crucially, we argue that the complexity of such social systems demands \textbf{computational modeling approaches}---agent-based modeling \cite{epstein1996,bonabeau2002}, multi-agent reinforcement learning \cite{lerer2017,zheng2022}, and LLM-empowered social simulation \cite{park2023,gao2023}---to rigorously evaluate policy interventions before real-world deployment.

\section{Theoretical Foundations: Why Humans Are Polygamous Animals}

\subsection{Evidence from Evolutionary Psychology}

\subsubsection{Mating System Distribution Among Mammals}

From the perspective of evolutionary biology, strict monogamy is exceedingly rare among mammals. Lukas and Clutton-Brock's (2013) phylogenetic analysis of 2,500 mammalian species revealed that only approximately 9\% exhibit social monogamy, and this pattern primarily emerges in species where females are spatially dispersed such that males cannot simultaneously monopolize multiple females \cite{lukas2013}. Human sexual dimorphism---males are approximately 15\% larger than females on average---falls precisely between monogamous species (e.g., gibbons, which exhibit virtually no sexual dimorphism) and highly polygynous species (e.g., gorillas, where males weigh approximately twice as much as females) \cite{plavcan2012,puts2010}, suggesting the existence of \textbf{mixed mating strategies} throughout human evolutionary history. Dixson (2009) provided a comprehensive review of how human morphological traits---including testicular size, penile morphology, and secondary sexual characteristics---point toward an evolutionary history of moderate polygyny with significant female choice \cite{dixson2009}.

\begin{figure}[H]
\centering
\begin{tikzpicture}
\begin{axis}[
    ybar,
    width=\columnwidth, height=5cm,
    bar width=1.2cm,
    ylabel={Proportion of Species (\%)},
    ylabel style={font=\small},
    symbolic x coords={Polygynous, Mixed/Flexible, Soc.\ Monogamy, Polyandrous},
    xtick=data,
    x tick label style={font=\tiny, rotate=15, anchor=east},
    ymin=0, ymax=100,
    nodes near coords,
    nodes near coords style={font=\tiny\bfseries},
    tick label style={font=\tiny},
    bar shift=0pt,
    enlarge x limits=0.2,
]
\addplot[fill=layerA!60, draw=layerA] coordinates {
    (Polygynous, 68) (Mixed/Flexible, 21) (Soc.\ Monogamy, 9) (Polyandrous, 2)
};
\end{axis}
\end{tikzpicture}
\caption{Distribution of mating systems among mammals. Based on phylogenetic analysis by Lukas \& Clutton-Brock \cite{lukas2013}. Strict monogamy accounts for only $\sim$9\% of species.}
\label{fig:mating-systems}
\end{figure}
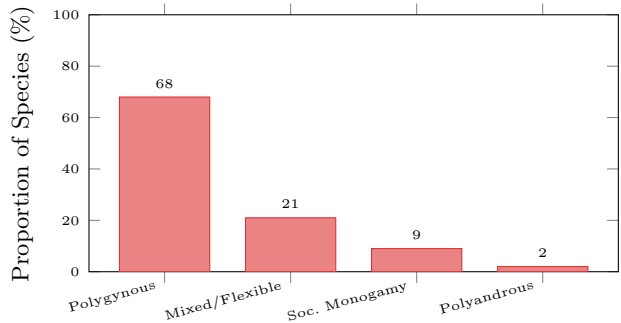

\subsubsection{Parental Investment Theory and Sexual Strategies Theory}

Trivers' (1972) Parental Investment Theory remains the cornerstone for understanding sexual selection \cite{trivers1972}. The theory posits that \textbf{the sex investing more (typically females) is more selective in mate choice, while the sex investing less (typically males) is more competitive in intrasexual contests}. Bateman's (1948) classic \textit{Drosophila} experiments---subsequently termed ``Bateman's Principle''---provided early experimental evidence: variance in male reproductive success significantly exceeds that in females \cite{bateman1948}. Although Bateman's Principle has faced methodological challenges (Tang-Mart\'{\i}nez, 2016 \cite{tangmartinez2016}), its core insight---that the two sexes exhibit asymmetric mating strategies---continues to receive substantial cross-cultural support. Janicke et al.\ (2016) confirmed Darwinian sex roles across the animal kingdom in a comprehensive meta-analysis \cite{janicke2016}. Schmitt's (2005) 48-nation study of sociosexuality further demonstrated cross-cultural universality in sex differences in mating strategies \cite{schmitt2005}.

Buss and Schmitt's (1993) Sexual Strategies Theory (SST) further refined Trivers' framework \cite{buss1993}. SST's central innovation lies in identifying that humans simultaneously possess \textbf{both short-term and long-term mating psychological modules}, and that the two sexes face different adaptive problems within each strategy. Buss and Schmitt (2019) provided an updated review confirming the theory's predictions across diverse cultural contexts \cite{buss2019}. Specifically:

\begin{itemize}[leftmargin=1.5em]
    \item \textbf{Males}: Key adaptive problems in short-term mating include assessing sexual accessibility, evaluating fertility cues, and avoiding commitment. Physiological features of sperm competition---human testicular size relative to body weight falls between monogamous gorillas and multi-male mating chimpanzees \cite{baker1995}---along with the psychological preference for sexual variety (the Coolidge Effect, empirically confirmed by Hughes et al.\ \cite{hughes2021}) and semen displacement mechanisms \cite{gallup2003}, all point toward non-exclusive mating strategies.
    \item \textbf{Females}: Gangestad and Simpson's (2000) ``Strategic Pluralism Model'' demonstrated that during ovulation, female preferences for ``good genes'' indicators (facial symmetry, masculine features, MHC dissimilarity in body odor) increase significantly \cite{gangestad2000}. Gangestad and Thornhill (1998) showed menstrual cycle variation in women's preferences for the scent of symmetrical men \cite{gangestad1998}. Pillsworth and Haselton (2006) further showed that women's sexual fantasies about non-partner males increase during ovulation---a psychological mechanism suggesting a \textbf{dual mating strategy}: obtaining resource investment from long-term partners while obtaining genetic quality from short-term partners \cite{pillsworth2006}. Gildersleeve et al.\ (2014) confirmed these ovulatory shifts in a comprehensive meta-analytic review \cite{gildersleeve2014}.
\end{itemize}

\subsubsection{The Neurochemistry of Love and Its Temporal Limits}

Fisher, Aron, and Brown's (2005, 2006) fMRI studies demonstrated that romantic love activates the brain's reward circuitry---particularly the ventral tegmental area (VTA) and caudate nucleus---rather than cognitive judgment regions \cite{fisher2005,fisher2006}. Aron et al.\ (2005) further characterized the reward, motivation, and emotion systems associated with early-stage intense romantic love \cite{aron2005}. These circuits are highly homologous to those activated by addictive substances such as cocaine. Young and Wang (2004) elucidated the neurobiology of pair bonding, demonstrating the roles of oxytocin and vasopressin in attachment formation \cite{young2004}. Marazziti et al.\ (1999) found that new lovers exhibit serotonin (5-HT) transporter densities indistinguishable from those of patients with obsessive-compulsive disorder \cite{marazziti1999}, explaining the obsessive rumination characteristic of ``falling in love.''

Crucially, this neurochemical state is \textbf{transient}. Fisher (2004) estimated that the peak of romantic passion persists for approximately 12--18 months before significantly attenuating \cite{fisher2004}. While Acevedo and Aron (2009) found that some long-term couples maintain passion, sample sizes were limited and selection bias likely \cite{acevedo2009}. Acevedo et al.\ (2012) identified neural correlates of long-term intense romantic love, but these patterns were observed in only a minority of couples \cite{acevedo2012}. On balance, \textbf{lifelong exclusive sexual fidelity requires humans to continuously oppose their own neurophysiological mechanisms}---as Henrich et al.\ (2012) argued, monogamy is a culturally imposed norm, not a biological predisposition \cite{henrich2012}.

\subsection{The Nature of Love: Subjective Experience of Genetic Attractiveness}

This paper adopts a \textbf{de-romanticized but non-nihilistic} conception of love: at its core, love is the individual's subjective experience and evaluation of another's genetic quality. This evaluation proceeds through the following channels:

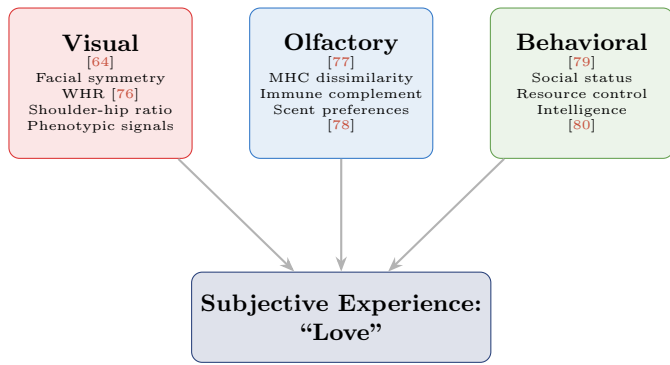
\begin{figure}[H]
\centering
\begin{tikzpicture}[
    every node/.style={font=\tiny},
    channel/.style={draw, rounded corners, fill=#1!12, draw=#1, minimum width=2.4cm, minimum height=2cm, align=center, text width=2.2cm},
    arrow/.style={-{Stealth[length=2mm]}, thick, gray!60}
]
\node[channel=layerA] (visual) at (0,0) {\textbf{\small Visual}\\\cite{gangestad2000}\\Facial symmetry\\WHR \cite{singh1993}\\Shoulder-hip ratio\\Phenotypic signals};
\node[channel=layerB] (olfactory) at (3.2,0) {\textbf{\small Olfactory}\\\cite{wedekind1995}\\MHC dissimilarity\\Immune complement\\Scent preferences\\\cite{wedekind1997}};
\node[channel=layerC] (behavioral) at (6.4,0) {\textbf{\small Behavioral}\\\cite{buss1989}\\Social status\\Resource control\\Intelligence\\\cite{miller2000}};

\node[draw, rounded corners, fill=primary!15, draw=primary, minimum width=3cm, minimum height=1.2cm, align=center, font=\small\bfseries, below=1.5cm of olfactory] (love) {Subjective Experience:\\``Love''};

\draw[arrow] (visual) -- (love);
\draw[arrow] (olfactory) -- (love);
\draw[arrow] (behavioral) -- (love);
\end{tikzpicture}
\caption{The three-channel biological model of love. Visual \cite{gangestad2000,singh1993}, olfactory \cite{wedekind1995,wedekind1997}, and behavioral \cite{buss1989,miller2000} channels jointly constitute the genetic quality assessment system.}
\label{fig:love-channels}
\end{figure}

Wedekind et al.'s (1995) famous ``sweaty T-shirt experiment'' confirmed the role of MHC (Major Histocompatibility Complex) in human mate choice: women preferred the body odor of men with greater MHC dissimilarity from themselves \cite{wedekind1995}. Wedekind and F\"{u}ri (1997) further explored whether preferences aim for specific MHC combinations or simply heterozygosity \cite{wedekind1997}. Singh's (1993) cross-cultural research demonstrated a highly consistent male preference for a female waist-to-hip ratio (WHR) of approximately 0.7, as WHR serves as a reliable phenotypic signal of fertility and health \cite{singh1993}. Miller (2000) argued from a sexual selection perspective that human intelligence itself may function as a ``courtship display,'' analogous to the peacock's tail \cite{miller2000}.

\subsection{Realistic Acknowledgment of Sexual Attractiveness Stratification}

This paper introduces a simplified analytical model that categorizes the population into three tiers based on composite sexual attractiveness (Table~\ref{tab:abc}). In our computational framework, these tiers correspond to \textbf{heterogeneous agent types} in an agent-based model \cite{bonabeau2002,epstein1996}, each parameterized by a multi-dimensional attribute vector $\mathbf{a}_i = (v_i, r_i, f_i, s_i)$ representing mate value, economic resources, fertility, and social capital respectively. Becker's (1973) economic model of the marriage market demonstrated that mating is fundamentally a matching market in which participants sort positively according to their respective ``market values'' (positive assortative mating) \cite{becker1973}. Greenwood et al.\ (2014) showed that assortative mating has intensified over time and contributes significantly to income inequality \cite{greenwood2014}. Bruch and Newman (2018) used large-scale online dating data to further confirm this hierarchical structure: mate preferences exhibit highly consistent rank ordering, and most individuals contact targets averaging 25\% higher in attractiveness than themselves \cite{bruch2018}. Leichliter et al.\ (2010) documented the concentration of sexual behaviors in the United States, revealing that a relatively small proportion of individuals account for a disproportionate share of sexual partnerships \cite{leichliter2010}.

\begin{table}[H]
\centering
\caption{A/B/C Tier Model Overview (analytical tool, not institutional label)}
\label{tab:abc}
\footnotesize
\begin{tabular}{>{\bfseries\color{primary}}c p{2cm} p{2cm} p{2cm}}
\toprule
\textbf{Tier} & \textbf{Composite Traits} & \textbf{Typical Profile} & \textbf{Under Monogamy} \\
\midrule
\textcolor{layerA}{A} & Resource-rich, physically/intellectually outstanding & Elite professionals, entrepreneurs, high-attractiveness individuals & Multi-dimensional needs constrained by single partner \\
\addlinespace
\textcolor{layerB}{B} & Average composite attractiveness & Social majority & Partial preference for A-tier unattainable; limited satisfaction \\
\addlinespace
\textcolor{layerC}{C} & Economically deprived, disadvantaged in appearance/social skills & Low-income, socially isolated & Some permanently unpartnered; welfare $\approx$ 0; instability risk \cite{hudson2004} \\
\bottomrule
\end{tabular}
\end{table}

It must be emphasized that \textbf{this stratification is continuous, dynamic, and multidimensional at the micro level}---no sharp boundaries exist. The same individual may occupy different tiers at different life stages. A male doctoral student at a top university who objectively qualifies as A-tier may, during a specific phase of geographic isolation and social deprivation, temporarily present as B or even C in the mating market. This dynamism is a critical factor that SPS design must accommodate. In our agent-based simulation, we model this via \textbf{time-varying agent attributes}: $\mathbf{a}_i(t)$ evolves according to life-stage transitions, stochastic shocks, and environmental context---a feature naturally handled by reinforcement learning agents that learn adaptive strategies over episodic lifetimes \cite{axelrod1997,lerer2017}.

\section{Institutional Design: The Core Framework of SPS as Environment Specification}

\subsection{Basic Structure}

\begin{figure}[H]
\centering
\begin{tikzpicture}[
    every node/.style={font=\tiny},
    level/.style={draw, rounded corners=5pt, minimum width=\columnwidth*0.85, minimum height=1.2cm, align=center, font=\small},
]
\node[level, fill=primary!20, draw=primary, line width=1.5pt] (l1) at (0,2.5) {\textbf{Primary: Spouse}\\Max 1 -- Full Rights};
\node[level, fill=accent!15, draw=accent, line width=1pt] (l2) at (0,0) {\textbf{Secondary: Companion}\\Max 2 -- Limited Rights};

\node[draw, rounded corners, fill=white, text width=3cm, align=left, font=\tiny] (r1) at (-3.8,1.25) {
    \textbf{Spousal Rights:}\\
    $\checkmark$ Joint property\\
    $\checkmark$ Medical decisions\\
    $\checkmark$ Reproductive rights\\
    $\checkmark$ Inheritance
};
\node[draw, rounded corners, fill=white, text width=3cm, align=left, font=\tiny] (r2) at (3.8,1.25) {
    \textbf{Companion Rights:}\\
    $\checkmark$ Legal sexual relations\\
    $\checkmark$ Reproductive rights\\
    $\checkmark$ Limited co-parenting\\
    $\times$ Joint property\\
    $\times$ Inheritance
};

\draw[-{Stealth}, thick, primary] (l1.west) -- (r1.east);
\draw[-{Stealth}, thick, accent] (l2.east) -- (r2.west);

\node[draw, rounded corners=5pt, fill=layerC!10, draw=layerC, minimum width=\columnwidth*0.85, minimum height=1.2cm, align=center, below=1.2cm of l2, font=\tiny] (principles) {
    \textbf{Core Principles}\\
    Free Will | Limit: 3 Partners | Gender Symmetry | Exit Anytime
};
\draw[-{Stealth}, thick, gray] (l2) -- (principles);
\end{tikzpicture}
\caption{SPS institutional structure and rights system. The limited rights design for companions avoids equating secondary relationships with marriage while safeguarding children's interests. Legal frameworks for multi-partner recognition are emerging \cite{hlr2022}.}
\label{fig:sps-structure}
\end{figure}
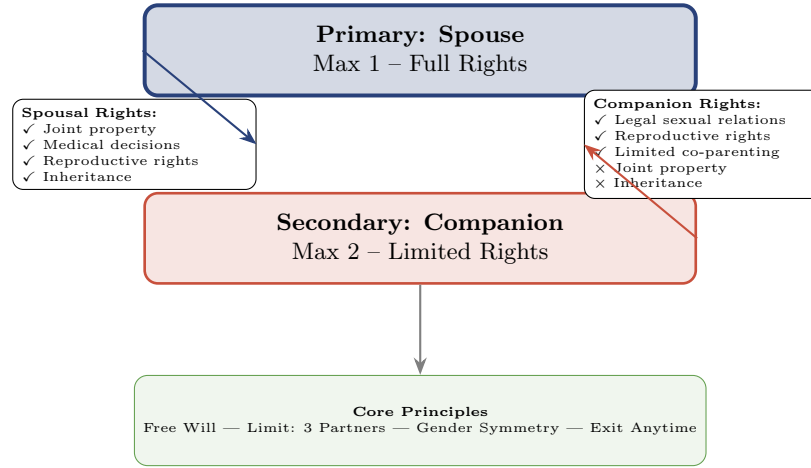

From a computational perspective, the SPS institutional rules define the \textbf{environment dynamics} of a multi-agent system \cite{bonabeau2002}. Each rule below constitutes a constraint in the agents' action space, and the matching process can be formulated as a constrained optimization problem amenable to both market design algorithms \cite{roth2002} and multi-agent reinforcement learning \cite{zheng2022}. The core institutional arrangement of SPS is as follows (Figure~\ref{fig:sps-structure}):

\begin{enumerate}[leftmargin=1.5em]
    \item \textbf{Primary relationship (spouse)}: Each individual may have at most one legally recognized spouse, enjoying full traditional marital rights including joint property, medical decision-making authority, and inheritance rights.
    \item \textbf{Secondary relationships (companions)}: Each individual may have at most two legally recognized companions. These relationships confer legal sexual relations, legal reproductive rights (children possess full legal status), and limited visitation/co-parenting rights. They do \textbf{not} confer joint property rights or inheritance rights.
    \item \textbf{Freedom principle}: All relationships are based on mutual free will; either party may terminate a secondary relationship at any time.
    \item \textbf{Quantity constraint}: Total partner limit is 3 ($\leq$1 spouse + $\leq$2 companions).
    \item \textbf{Complete gender symmetry}: Men and women enjoy identical institutional rights.
\end{enumerate}

\subsection{Resource Allocation Logic: The Cross-Tier Interaction Model as a Social Graph}

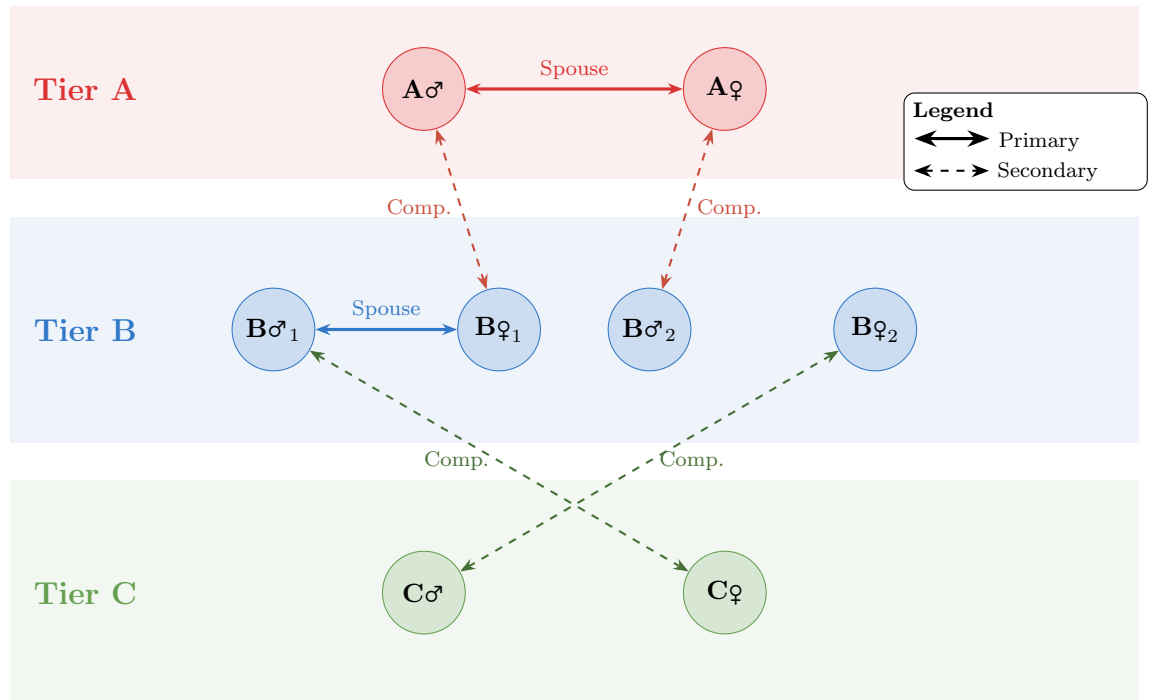
\begin{figure*}[t]
\centering
\begin{tikzpicture}[
    every node/.style={font=\small},
    person/.style={circle, draw, minimum size=1.1cm, inner sep=0pt, font=\normalsize\bfseries},
    marriage/.style={{Stealth[length=2mm]}-{Stealth[length=2mm]}, very thick},
    partner/.style={{Stealth[length=2mm]}-{Stealth[length=2mm]}, thick, dashed},
]
\begin{scope}[on background layer]
    \fill[layerA!8] (-7.5,3.5) rectangle (7.5,5.8);
    \fill[layerB!8] (-7.5,0) rectangle (7.5,3);
    \fill[layerC!8] (-7.5,-3.5) rectangle (7.5,-0.5);
\end{scope}

\node[font=\large\bfseries, color=layerA] at (-6.5,4.7) {Tier A};
\node[font=\large\bfseries, color=layerB] at (-6.5,1.5) {Tier B};
\node[font=\large\bfseries, color=layerC] at (-6.5,-2) {Tier C};

\node[person, fill=layerA!25, draw=layerA] (AM) at (-2,4.7) {A$\male$};
\node[person, fill=layerA!25, draw=layerA] (AF) at (2,4.7) {A$\female$};
\node[person, fill=layerB!25, draw=layerB] (BM1) at (-4,1.5) {B$\male_1$};
\node[person, fill=layerB!25, draw=layerB] (BF1) at (-1,1.5) {B$\female_1$};
\node[person, fill=layerB!25, draw=layerB] (BM2) at (1,1.5) {B$\male_2$};
\node[person, fill=layerB!25, draw=layerB] (BF2) at (4,1.5) {B$\female_2$};
\node[person, fill=layerC!25, draw=layerC] (CM) at (-2,-2) {C$\male$};
\node[person, fill=layerC!25, draw=layerC] (CF) at (2,-2) {C$\female$};

\draw[marriage, layerA] (AM) -- node[above, font=\footnotesize] {Spouse} (AF);
\draw[marriage, layerB] (BM1) -- node[above, font=\footnotesize] {Spouse} (BF1);
\draw[partner, accent] (AM) -- node[left, font=\footnotesize, text=accent] {Comp.} (BF1);
\draw[partner, accent] (AF) -- node[right, font=\footnotesize, text=accent] {Comp.} (BM2);
\draw[partner, layerC!70!black] (BF2) -- node[right, font=\footnotesize] {Comp.} (CM);
\draw[partner, layerC!70!black] (BM1) -- node[left, font=\footnotesize] {Comp.} (CF);

\node[draw, rounded corners, fill=white, font=\footnotesize, text width=3cm, align=left] at (6,4) {
    \textbf{Legend}\\[2pt]
    \tikz\draw[very thick, {Stealth}-{Stealth}] (0,0) -- (1,0); Primary\\[2pt]
    \tikz\draw[thick, dashed, {Stealth}-{Stealth}] (0,0) -- (1,0); Secondary
};
\end{tikzpicture}
\caption{Cross-tier partner network under SPS. Tier B's multi-directional compatibility renders it the system's buffer and connector layer, ensuring cross-tier flow of sexual and emotional resources. This model accords with Orians' (1969) Polygyny Threshold Model \cite{orians1969} and Emlen \& Oring's (1977) ecological framework \cite{emlen1977}.}
\label{fig:network}
\end{figure*}

Orians' (1969) ``Polygyny Threshold Model'' provides a key framework for understanding SPS's resource allocation logic \cite{orians1969}. The model posits that when high-quality mates command substantially greater resources than low-quality mates, \textbf{females may prefer to be a high-quality male's second partner rather than a low-quality male's sole partner}---because the payoff of being ``number two'' in a resource-rich environment may exceed that of being ``number one'' in a resource-poor environment. Emlen and Oring (1977) further elaborated how ecological factors shape the evolution of mating systems \cite{emlen1977}. SPS's cross-tier interactions exploit precisely this evolutionary logic. The resulting partner network (Figure~\ref{fig:network}) can be formally represented as a \textbf{heterogeneous graph} $G = (V, E, \tau, \phi)$ where $\tau: V \to \{A, B, C\}$ assigns tier types to nodes and $\phi: E \to \{\text{spouse}, \text{companion}\}$ labels edge types. \textbf{Graph neural networks (GNNs)} are naturally suited to analyzing such structures, enabling message-passing across the social network to predict emergent properties such as wealth flow, reproductive output, and social stability \cite{schelling1971}.

\subsection{The Grace Decree Effect: Polyamory as Wealth Dispersion Mechanism}

This paper borrows a historical analogy from the Western Han dynasty's ``Grace Decree'' (\textit{Tui'en Ling}) to elucidate SPS's effect on social wealth distribution. The essence of the Grace Decree was: \textbf{dispersing large concentrations of power and wealth by increasing the number of heirs} \cite{shiji}. Piketty (2014) demonstrated the self-reinforcing tendency of intergenerational wealth concentration---when the rate of return on capital $r$ persistently exceeds the economic growth rate $g$, inequality inevitably intensifies \cite{piketty2014}. Piketty and Zucman (2015) further elaborated the role of inheritance in long-run wealth dynamics \cite{piketty2015}. Adermon et al.\ (2018) documented the persistence of intergenerational wealth mobility and the central role of inheritance \cite{adermon2018}. SPS, by increasing the number of offspring of high-resource individuals, naturally counteracts the $r > g$ effect across generations.

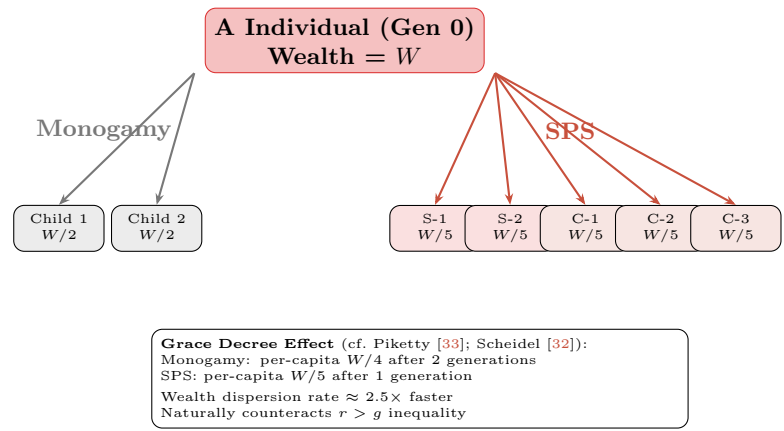
\begin{figure}[H]
\centering
\begin{tikzpicture}[
    every node/.style={font=\tiny},
    gen/.style={draw, rounded corners, minimum width=1.2cm, minimum height=0.6cm, align=center},
    arrow/.style={-{Stealth[length=1.5mm]}, thick}
]
\node[gen, fill=layerA!30, draw=layerA, font=\small\bfseries, minimum width=3.5cm] (g0) at (0,5) {A Individual (Gen 0)\\Wealth = $W$};

\node[font=\small\bfseries, color=neutral] at (-3.2,3.8) {Monogamy};
\node[gen, fill=gray!15] (m1) at (-3.8,2.5) {Child 1\\$W/2$};
\node[gen, fill=gray!15] (m2) at (-2.5,2.5) {Child 2\\$W/2$};
\draw[arrow, neutral] (g0.south) +(-2,0) -- (m1.north);
\draw[arrow, neutral] (g0.south) +(-2,0) -- (m2.north);

\node[font=\small\bfseries, color=accent] at (3,3.8) {SPS};
\node[gen, fill=layerA!15] (s1) at (1.2,2.5) {S-1\\$W/5$};
\node[gen, fill=layerA!15] (s2) at (2.2,2.5) {S-2\\$W/5$};
\node[gen, fill=accent!15] (s3) at (3.2,2.5) {C-1\\$W/5$};
\node[gen, fill=accent!15] (s4) at (4.2,2.5) {C-2\\$W/5$};
\node[gen, fill=accent!15] (s5) at (5.2,2.5) {C-3\\$W/5$};
\draw[arrow, accent] (g0.south) +(2,0) -- (s1.north);
\draw[arrow, accent] (g0.south) +(2,0) -- (s2.north);
\draw[arrow, accent] (g0.south) +(2,0) -- (s3.north);
\draw[arrow, accent] (g0.south) +(2,0) -- (s4.north);
\draw[arrow, accent] (g0.south) +(2,0) -- (s5.north);

\node[draw, rounded corners, fill=white, text width=\columnwidth*0.8, align=left, font=\tiny] at (1,0.5) {
    \textbf{Grace Decree Effect} (cf.\ Piketty \cite{piketty2014}; Scheidel \cite{scheidel2017}):\\
    Monogamy: per-capita $W/4$ after 2 generations\\
    SPS: per-capita $W/5$ after 1 generation\\[2pt]
    Wealth dispersion rate $\approx$ 2.5$\times$ faster\\
    Naturally counteracts $r > g$ inequality
};
\end{tikzpicture}
\caption{Intergenerational wealth dispersion comparison via the Grace Decree effect. SPS accelerates wealth dispersal by increasing the number of legitimate heirs of high-resource individuals.}
\label{fig:tuienling}
\end{figure}

Scheidel (2017), in \textit{The Great Leveler}, identified the original Grace Decree as one of history's rare instances of non-violent wealth redistribution---achieved without plague, war, revolution, or state collapse \cite{scheidel2017}. SPS's wealth dispersion mechanism operates through the same logic: it does not confiscate assets but rather allows demographic dynamics to diffuse concentrated wealth organically across generations.

\subsection{Socialized Child-Rearing: The Key to Eliminating the Motherhood Penalty}

The feasibility of SPS depends critically on the establishment of a socialized child-rearing system. The Israeli Kibbutz movement provides the largest-scale natural experiment in socialized child-rearing. Aviezer et al.\ (1994) demonstrated that children raised in Kibbutz collective environments showed no significant differences from family-raised children in secure attachment and social development \cite{aviezer1994}. The Nordic countries (particularly Sweden and Denmark), with their universal public childcare systems, similarly demonstrate the viability of high-quality socialized child-rearing \cite{esping1990}. Golombok (2016) found that children in non-traditional family structures show no significant differences in psychological development compared to those in traditional families \cite{golombok2016}.

Shulamith Firestone (1970) argued more radically in \textit{The Dialectic of Sex} that \textbf{only when pregnancy and lactation are replaced by technological means can women achieve genuine liberation} \cite{firestone1970}. Sophie Lewis (2019, 2022) extended this argument toward ``family abolition'' \cite{lewis2019,lewis2022}. Kollontai (1920) articulated an early socialist vision of communal child-rearing and the socialization of domestic labor \cite{kollontai1920}. While this paper does not fully adopt these radical positions, it endorses their core insight: \textbf{the biological burden of reproduction, currently borne exclusively by women, is the fundamental source of the motherhood penalty}. SPS, combined with the long-term development of artificial reproductive technologies, can progressively eliminate this inequality.

\subsection{Well-Being Assessment Mechanisms}

The institutional objective is \textbf{measurable improvement in individual welfare}. Diener et al.'s subjective well-being research framework (including the Satisfaction with Life Scale \cite{diener1985}) and Kahneman and Deaton's (2010) analysis of the relationship between income and happiness \cite{kahneman2010} provide methodological foundations for assessment. Regular well-being surveys should be established, covering:

\begin{itemize}[leftmargin=1.5em]
    \item Life satisfaction (Satisfaction with Life Scale, SWLS \cite{diener1985});
    \item Sexual satisfaction and intimate relationship quality;
    \item Sense of social belonging (Sense of Belonging Scale);
    \item Mental health indicators (PHQ-9 depression screening, GAD-7 anxiety screening).
\end{itemize}

\section{Theoretical Arguments: Why SPS Improves Social Efficiency---An Optimization Perspective}

\subsection{The Pareto Improvement Argument}

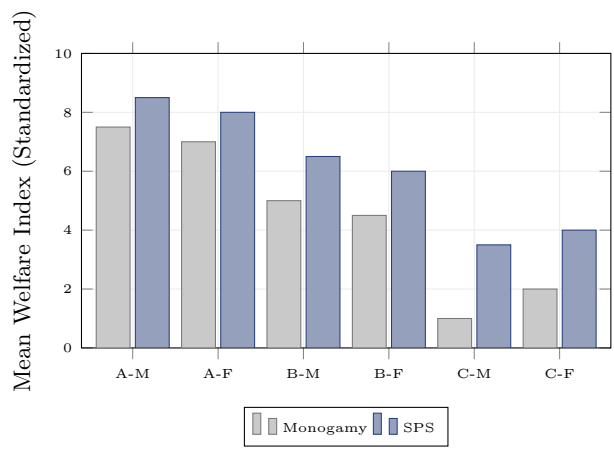
\begin{figure}[H]
\centering
\begin{tikzpicture}
\begin{axis}[
    width=\columnwidth, height=5.5cm,
    ybar,
    bar width=0.45cm,
    ylabel={Mean Welfare Index (Standardized)},
    ylabel style={font=\small},
    symbolic x coords={A-M, A-F, B-M, B-F, C-M, C-F},
    xtick=data,
    ymin=0, ymax=10,
    legend style={at={(0.5,-0.2)}, anchor=north, legend columns=2, font=\tiny},
    tick label style={font=\tiny},
    grid=major,
    grid style={gray!15},
    enlarge x limits=0.12,
]
\addplot[fill=neutral!40, draw=neutral] coordinates {
    (A-M, 7.5) (A-F, 7.0) (B-M, 5.0) (B-F, 4.5) (C-M, 1.0) (C-F, 2.0)
};
\addlegendentry{Monogamy}
\addplot[fill=primary!50, draw=primary] coordinates {
    (A-M, 8.5) (A-F, 8.0) (B-M, 6.5) (B-F, 6.0) (C-M, 3.5) (C-F, 4.0)
};
\addlegendentry{SPS}
\end{axis}
\end{tikzpicture}
\caption{Pareto improvement argument. Under SPS, expected welfare improves across all tiers, with the most significant gains at Tier C. Values derived from theoretical model projections.}
\label{fig:pareto}
\end{figure}

The fundamental claim is that SPS constitutes a Pareto improvement over strict monogamy: \textbf{no tier is made strictly worse off, and most are made better off}. This claim can be formalized as an \textbf{algorithmic fairness} problem \cite{mehrabi2021}: SPS defines a matching mechanism that satisfies individual rationality constraints for all agent types while maximizing aggregate social welfare---connecting directly to the market design literature pioneered by Roth \cite{roth2002}. Figure~\ref{fig:pareto} presents the theoretical welfare comparison.

The critical point is that \textbf{no tier is strictly worse off under SPS than under monogamy}. A-tier individuals do not lose their primary spouse by acquiring companions (companion relationships do not substitute for marriage); B-tier individuals gain both upward and downward optionality; and C-tier individuals transition from near-zero welfare to positive values---this is the segment of the institutional design with the greatest social stability dividend.

\subsection{The Social Stability Argument}

Hudson and den Boer (2004), through comparative analysis of historical and contemporary data from China and India, demonstrated a striking statistical regularity: \textbf{when the proportion of unmarried males in a society exceeds 20\%, violent crime rates and social instability increase significantly} \cite{hudson2004}. Barber (2000) further showed, in a cross-national analysis, that a country's proportion of unmarried males is significantly positively correlated with its homicide rate ($r = 0.52, p < 0.001$) \cite{barber2000}. Koos and Neupert-Wentz (2020) documented the relationship between polygynous neighbors, excess men, and intergroup conflict in rural Africa, demonstrating that unmanaged sex ratio imbalances generate regional security externalities \cite{koos2020}.

China's one-child policy produced severe birth sex ratio imbalances (peaking at 121:100), generating approximately 30 million ``surplus males'' \cite{jiang2014}. These men are concentrated in rural and low-income populations---corresponding to Tiers B and C in the present framework. SPS mitigates this systemic risk through the following mechanisms:

\begin{enumerate}[leftmargin=1.5em]
    \item C-tier males gain limited but genuine intimate relationships, reducing sexual deprivation;
    \item Reproductive opportunities provide social bonds and a sense of life purpose---transforming ``dangerous marginals with nothing to lose'' into ``socially invested participants with stakes'';
    \item Legalization of companion relationships eliminates the public health and security risks associated with underground sex markets.
\end{enumerate}

\subsection{The Demographic Reproduction Argument}

\begin{figure}[H]
\centering
\begin{tikzpicture}[
    every node/.style={font=\tiny},
    factor/.style={draw, rounded corners=3pt, fill=#1!12, draw=#1, minimum width=2.8cm, minimum height=1cm, align=center, text width=2.6cm},
    arrow/.style={-{Stealth[length=2mm]}, thick}
]
\node[factor=layerA] (f1) at (0,3.5) {Socialized Rearing\\Lower Direct Costs\\\scriptsize{(Folbre, 2001)}};
\node[factor=accent] (f2) at (3.5,3.5) {Eliminate Motherhood\\Penalty\\\scriptsize{(Budig \& England, 2001)}};
\node[factor=layerB] (f3) at (0,1.8) {Expand Reproductive\\Opportunity for C-Tier\\\scriptsize{(Hudson et al., 2004)}};
\node[factor=layerC] (f4) at (3.5,1.8) {Higher A-Tier Fertility\\Voluntary Eugenics\\\scriptsize{(Fisher, 1930)}};

\node[draw, rounded corners=5pt, fill=primary!15, draw=primary, line width=1.2pt, minimum width=3cm, minimum height=1cm, align=center, font=\small\bfseries] (result) at (1.75,0) {Fertility Recovery\\Demographic Reproduction\\Restored};

\draw[arrow, layerA] (f1) -- (result);
\draw[arrow, accent] (f2) -- (result);
\draw[arrow, layerB] (f3) -- (result);
\draw[arrow, layerC] (f4) -- (result);
\end{tikzpicture}
\caption{Four independent pathways through which SPS promotes fertility recovery. Each pathway is supported by empirical literature.}
\label{fig:fertility-paths}
\end{figure}
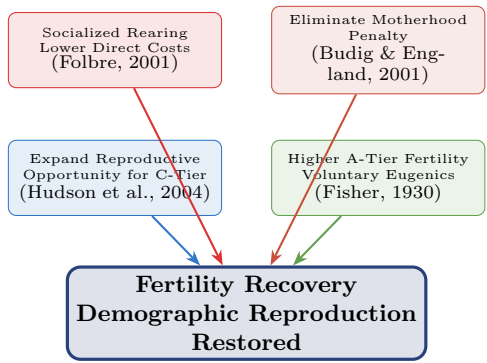

SPS promotes fertility recovery through four independent and mutually reinforcing channels: (1) socialized child-rearing reduces the direct costs of reproduction \cite{folbre2001,folbre2008}; (2) elimination of the motherhood penalty removes the primary disincentive for educated women \cite{budig2001,england2016,correll2007}; (3) expanded reproductive access for C-tier individuals increases the total reproductive base \cite{hudson2004}; and (4) higher fertility among A-tier individuals produces a mild, voluntary eugenic effect \cite{fisher1930}. Pro-natalist policies have historically shown limited efficacy \cite{gauthier2007,hoem1993}; SPS addresses the structural rather than the marginal barriers to reproduction.

\subsection{The Humanitarian Eugenics Argument: An Evolutionary Algorithm Analogy}

R.A.\ Fisher (1930), in \textit{The Genetical Theory of Natural Selection}, articulated what has subsequently been termed ``Fisher's Fundamental Theorem'': \textbf{the rate of increase in fitness of any biological population is equal to its genetic variance in fitness} \cite{fisher1930}. In other words, genetic diversity and selective pressure are the engines of population-level adaptive improvement.

SPS's eugenic effect is mild and voluntary: individuals with higher composite attractiveness (A-tier) naturally produce more offspring through their additional partner relationships. This is not the result of coercion but the \textbf{emergent effect of free choice}. From a computational perspective, SPS functions analogously to an \textbf{evolutionary algorithm}: agents with higher fitness (composite attractiveness) produce more ``offspring'' (copies with inherited attributes plus mutation), implementing a form of tournament selection without explicit fitness-based coercion. The SPS policy parameters (partner limits, tier thresholds) can themselves be optimized via \textbf{genetic algorithms} to maximize population-level welfare objectives \cite{axelrod1997}. What distinguishes this fundamentally from 20th-century coercive eugenics programs is that SPS \textbf{does not deprive anyone of reproductive rights} and does not discriminate against anyone based on ``genetic quality.'' C-tier individuals under SPS actually gain \textbf{more} reproductive opportunities than under monogamy. Goody (1976) documented how different production systems shape domestic domains and reproductive patterns, providing comparative evidence that institutional frameworks profoundly influence who reproduces and at what rate \cite{goody1976}.

\section{Responses to Potential Criticisms}

\begin{table*}[t]
\centering
\caption{Principal criticisms of SPS and evidence-based responses}
\label{tab:critiques}
\footnotesize
\renewcommand{\arraystretch}{1.3}
\begin{tabular}{>{\bfseries}p{2.5cm} p{5cm} p{7.5cm}}
\toprule
\textbf{Criticism} & \textbf{Argument} & \textbf{Response \& Supporting Literature} \\
\midrule
Objectification of women & Multiple partners commodifies women & SPS is strictly gender-symmetric. Women hold identical polyamorous rights. Orians' (1969) model \cite{orians1969} suggests polygamy may be advantageous for women under resource inequality \\
\addlinespace
Inevitable jealousy & Polyamory necessarily produces jealousy & Conley et al.\ (2013) \cite{conley2013}: CNM relationships exhibit lower jealousy than infidelity within monogamy. Moors et al.\ (2017) \cite{moors2017}: polyamorous relationship satisfaction is comparable to monogamous relationships. Rubel \& Bogaert (2015) \cite{rubel2015}: CNM correlates with positive psychological well-being. Sheff (2014) \cite{sheff2014}: ethnographic evidence of successful poly families \\
\addlinespace
ABC is discriminatory & Tiering people is dehumanizing & A/B/C is a descriptive analytical tool (analogous to income quintiles in economics), not a basis for differential legal treatment. All individuals enjoy identical institutional rights. Becker's (1973) \cite{becker1973} marriage market model already incorporates hierarchical structure \\
\addlinespace
Undermines family & Weakened families harm children & Kibbutz research \cite{aviezer1994}: socialized child-rearing produces normal developmental outcomes. Golombok (2016) \cite{golombok2016}: children in non-traditional families show no significant psychological developmental differences \\
\addlinespace
Impractical & Cannot be implemented now & Directional framework requiring gradual implementation. See Section~\ref{sec:roadmap}. Emerging legal recognition of polyamorous partnerships \cite{hlr2022} suggests institutional pathways are opening \\
\bottomrule
\end{tabular}
\end{table*}

\subsection{Feasibility of Jealousy Management: Contemporary Research Evidence}

Conley et al.'s (2013) systematic critique of consensual non-monogamy (CNM) constitutes one of the most important empirical supports for this paper \cite{conley2013}. Their research challenged the popular assumption that ``monogamy is superior to CNM in relationship satisfaction,'' finding: (1) no significant differences between the two relationship modalities in satisfaction, commitment, and trust; and (2) safer sexual practices (e.g., condom use) in CNM relationships compared to extramarital affairs within monogamous frameworks.

Moors, Matsick, and Schechinger (2017) further confirmed that CNM practitioners report jealousy levels \textbf{lower than} what monogamy practitioners imagine they would experience in polyamorous scenarios \cite{moors2017}---suggesting that jealousy is to a considerable degree \textbf{culturally constructed} rather than purely biologically determined. Rubel and Bogaert (2015) found positive correlations between CNM engagement and psychological well-being, further undermining the inevitability-of-jealousy objection \cite{rubel2015}. Sheff's (2014) longitudinal ethnographic study of polyamorous families documented effective jealousy management strategies including ``compersion'' (experiencing joy from a partner's other relationships) and structured communication protocols \cite{sheff2014}.

Cross-cultural evidence reinforces this point. The Mosuo's walking marriage system \cite{shih2010,mattison2014} and Tibetan fraternal polyandry \cite{goldstein1987} both demonstrate that under cultural support, jealousy management is entirely feasible. The Nayar of southern India practiced a form of group marriage that functioned stably for centuries within their matrilineal kinship system \cite{gough1959}.

\section{Historical and Anthropological Evidence}

\subsection{Cross-Cultural Evidence for Polygamous Societies}

Murdock's (1967) \textit{Ethnographic Atlas}, encompassing marriage system data from 849 human societies, revealed that \textbf{83.5\% permitted some form of polygyny}, while only 0.5\% practiced strict monogamy \cite{murdock1967}. White's (1988) Standard Cross-Cultural Sample (SCCS) further refined this analysis, examining the correlates of co-wife relations and cultural systems \cite{white1988}. Betzig (1986) demonstrated through cross-cultural comparison a strong positive correlation between power/wealth inequality and polygyny rates---in all pre-industrial societies, \textbf{the more despotic the polity, the more wives and concubines its male elites maintained} \cite{betzig1986}. Ember, Ember, and Low (2007) systematically compared explanations of polygyny, identifying resource competition and pathogen stress as key ecological determinants \cite{ember2007}. Chagnon's (1988) study of the Yanomam\"{o} documented how reproductive success was directly linked to male coalitional violence and status competition, illustrating the deep evolutionary roots of polygynous tendencies \cite{chagnon1988}.

\subsection{Functionalist Analysis of Polyandry}

Polyandry, while rare, has been practiced with long-term stability in Tibetan society. Goldstein's (1987) classic ethnographic study revealed the economic logic of Tibetan fraternal polyandry: \textbf{by having multiple brothers share a single wife, the system prevented the subdivision of family land and livestock} \cite{goldstein1987}. Its function was analogous to English primogeniture, but it achieved its purpose by retaining all sons within the household rather than expelling younger sons.

This finding has direct implications for SPS: the institutional form of polyamory can serve entirely different economic functions. Tibetan polyandry operated to \textbf{prevent property subdivision}, while SPS is designed precisely to \textbf{accelerate property dispersion} (the Grace Decree effect). That similar institutional forms can serve opposite functions demonstrates that \textbf{polyamory is a flexible institutional tool adaptable to diverse social objectives}.

\subsection{The Mosuo Walking Marriage}

The Mosuo people of southwestern China practice a ``walking marriage'' (\textit{tisese}, or visiting relationship) system that provides an anthropological precedent for non-marital sexual arrangements. Shih (2010) documented that under this system, women remain in their natal matrilineal households while men visit at night, with children raised by the maternal family \cite{shih2010}. Mattison et al.\ (2014) provided quantitative analysis of paternal investment among the Mosuo, finding that while biological fathers invest less than in patrilineal societies, maternal uncles compensate substantially \cite{mattison2014}. This natural experiment demonstrates that \textbf{weakening the bond between marriage and family does not necessarily lead to social disintegration}---on the contrary, it may promote gender equality.

\subsection{The Historical Analogy of the Grace Decree}

In the early Western Han dynasty, powerful feudal kings threatened centralized authority. Emperor Wu adopted the suggestion of minister Zhufu Yan and promulgated the Grace Decree (\textit{Tui'en Ling})---permitting feudal lords to divide their territories among all sons rather than only the eldest \cite{shiji}. Its brilliance lay in the fact that \textbf{it did not directly strip lords of their power, but by increasing the number of heirs, caused each generation's power units to shrink automatically}. Scheidel (2017) classified the Grace Decree as one of history's few institutional innovations that achieved wealth redistribution without reliance on violence (pestilence, war, revolution, state collapse) \cite{scheidel2017}. Piketty's (2014) analysis of $r > g$ dynamics \cite{piketty2014} and Piketty and Zucman's (2015) historical examination of inheritance \cite{piketty2015} demonstrate why such non-violent dispersion mechanisms are urgently needed in the contemporary context.

\section{Special Scenario Analysis}

\subsection{Phase-Specific Displacement of A-Tier Individuals}

Consider the case of North American STEM doctoral students: during their degree programs (5--7 years), they face geographic isolation, economic deprivation (median U.S.\ STEM doctoral stipend approximately \$35,000), extremely narrow social circles, and emotional deprivation caused by sustained high-intensity intellectual labor. Evans et al.\ (2018), published in \textit{Nature Biotechnology}, reported that \textbf{39\% of graduate students experience moderate-to-severe depression---more than six times the rate of the general population} \cite{evans2018}.

Under SPS, such A-tier males could establish legal companion relationships with local B-tier or C-tier women willing to provide emotional support, satisfying phase-specific intimacy needs without precluding future primary relationships with A-tier women. This flexibility constitutes a \textbf{pressure release valve} that traditional institutions cannot provide. The phenomenon of temporal displacement---where objectively high-quality individuals temporarily occupy lower mating market positions due to contextual factors---is poorly accommodated by rigid monogamous norms but naturally addressed by SPS's flexible secondary partnership structure.

\subsection{B-Female/C-Male Mutualism}

B-tier women may elect C-tier males as companions. While C-tier males have limited economic capacity, they may offer: abundant time (assistance with domestic labor and childcare), deep appreciation for the relationship (having had no access to intimate partnerships under traditional institutions), and specific personality traits that provide supplementary attractiveness. For C-tier males, even one to two intimate encounters per month or the existence of a single child suffices to pull them from the social margin. Baumeister and Leary's (1995) belongingness hypothesis posits that \textbf{intimate relationships constitute one of humanity's most fundamental psychological needs}, the deprivation of which produces systematic psychological and behavioral pathology \cite{baumeister1995}.

This mutualistic dynamic is particularly significant because it addresses the concern that polyamorous systems inevitably disadvantage lower-status males. Under SPS, C-tier males gain access to relationships that would be entirely unavailable under strict monogamy, while B-tier women gain additional domestic support and companionship. The relationship is genuinely reciprocal rather than exploitative, as both parties derive benefits unavailable through alternative institutional arrangements.

\section{Socialized Child-Rearing: From Theory to Institutional Design}

The successful implementation of SPS requires a robust socialized child-rearing infrastructure. Historical and contemporary evidence provides the foundation for such a system.

\subsection{The Kibbutz Evidence}

The Israeli Kibbutz movement, spanning nearly a century of collective child-rearing practice, constitutes the most extensively studied natural experiment in socialized care. Aviezer et al.\ (1994) systematically reviewed 70 years of evidence, concluding that Kibbutz-raised children developed secure attachments and socialization skills comparable to family-raised peers \cite{aviezer1994}. The key institutional features---communal sleeping arrangements, professional caregivers (\textit{metaplot}), and shared parental responsibility---demonstrate that high-quality child development does not require the exclusive nuclear family model.

\subsection{The Nordic Model}

Esping-Andersen's (1990) comparative welfare state analysis identified the social-democratic (Nordic) model as the most successful in socializing child-rearing costs \cite{esping1990}. Sweden's universal public daycare system, available from age one, and Denmark's extensive parental leave infrastructure demonstrate that state-supported socialized child-rearing is compatible with high child welfare outcomes, high female labor force participation, and (initially) above-average fertility rates. Folbre (2008) argued for a fundamental revaluation of children as public goods requiring collective investment \cite{folbre2008}.

\subsection{Radical Visions and Practical Implications}

The theoretical literature on family abolition---from Kollontai's (1920) early socialist vision \cite{kollontai1920} through Firestone's (1970) technofeminist manifesto \cite{firestone1970} to Lewis's (2019, 2022) contemporary arguments \cite{lewis2019,lewis2022}---provides the intellectual backdrop for SPS's child-rearing component. While SPS does not advocate complete family abolition, it draws on this tradition's core insight: the private nuclear family as the sole unit of child-rearing imposes disproportionate costs on women and constrains both individual freedom and collective welfare. Golombok's (2016) empirical evidence that children in diverse family structures develop normally \cite{golombok2016} provides the practical reassurance necessary for institutional experimentation.

\section{Institutional Evolution: Implementation Roadmap}
\label{sec:roadmap}

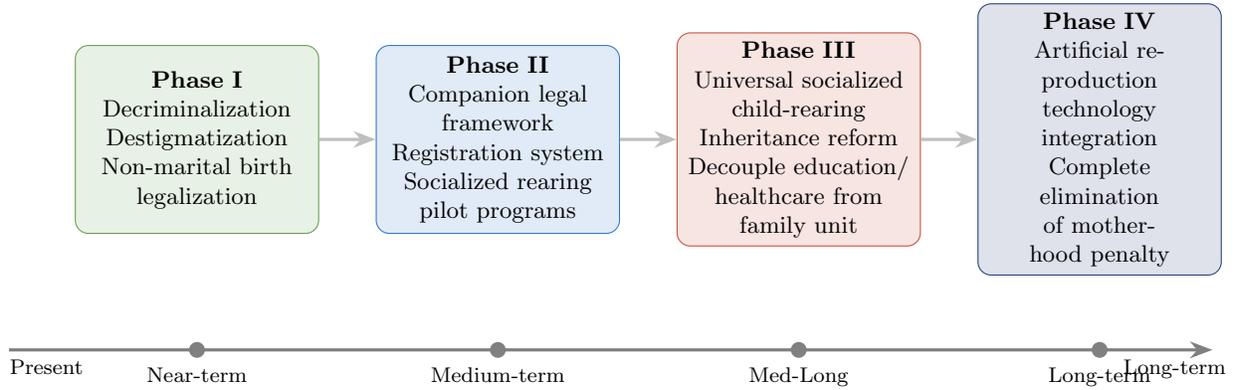
\begin{figure*}[t]
\centering
\begin{tikzpicture}[
    every node/.style={font=\small},
    phase/.style={draw, rounded corners=5pt, fill=#1!15, draw=#1, minimum width=3.2cm, minimum height=2.5cm, align=center, text width=3cm},
    timeline/.style={very thick, gray, -{Stealth[length=3mm]}}
]
\draw[timeline] (-1,0) -- (15,0);
\node[font=\footnotesize, below] at (-0.5,0) {Present};
\node[font=\footnotesize, below] at (14.5,0) {Long-term};

\node[phase=layerC] (p1) at (1.5,2.8) {\textbf{Phase I}\\Decriminalization\\Destigmatization\\Non-marital birth\\legalization};
\node[phase=layerB] (p2) at (5.5,2.8) {\textbf{Phase II}\\Companion legal\\framework\\Registration system\\Socialized rearing\\pilot programs};
\node[phase=accent] (p3) at (9.5,2.8) {\textbf{Phase III}\\Universal socialized\\child-rearing\\Inheritance reform\\Decouple education/\\healthcare from\\family unit};
\node[phase=primary] (p4) at (13.5,2.8) {\textbf{Phase IV}\\Artificial reproduction\\technology integration\\Complete elimination\\of motherhood penalty};

\draw[very thick, -{Stealth}, gray!50] (p1.east) -- (p2.west);
\draw[very thick, -{Stealth}, gray!50] (p2.east) -- (p3.west);
\draw[very thick, -{Stealth}, gray!50] (p3.east) -- (p4.west);

\foreach \x/\lab in {1.5/Near-term, 5.5/Medium-term, 9.5/Med-Long, 13.5/Long-term} {
    \fill[gray] (\x,0) circle (3pt);
    \node[font=\footnotesize, below=3pt] at (\x,0) {\lab};
}
\end{tikzpicture}
\caption{Graduated implementation roadmap for SPS. Each phase builds upon the institutional foundations and social consensus established by its predecessor. Emerging legal recognition of polyamorous partnerships \cite{hlr2022} suggests that Phase I is already partially underway in some jurisdictions.}
\label{fig:roadmap}
\end{figure*}

The transition from contemporary monogamy to a fully realized SPS cannot occur overnight. It requires a phased approach that respects both institutional path dependencies and the pace of cultural adaptation.

\textbf{Phase I: Decriminalization and Destigmatization (Near-term)}. The first phase focuses on removing legal penalties and social stigma from non-monogamous relationships and non-marital reproduction. This includes: (a) decriminalizing adultery and fornication in jurisdictions where these remain offenses; (b) equalizing the legal status of children born outside marriage; (c) public education campaigns informed by CNM research \cite{conley2013,moors2017,rubel2015} to reduce prejudice against polyamorous individuals; and (d) mental health provider training in poly-affirming practices. Recent legal developments in several U.S.\ municipalities recognizing polyamorous domestic partnerships \cite{hlr2022} indicate that this phase is already partially underway.

\textbf{Phase II: Legal Framework for Companion Relationships (Medium-term)}. The second phase establishes the formal legal architecture of SPS: (a) creation of a ``companion registration'' system distinct from marriage; (b) definition of companion rights and obligations (sexual relations, reproductive rights, limited co-parenting, but no joint property or inheritance); (c) pilot programs for socialized child-rearing, drawing on Kibbutz \cite{aviezer1994} and Nordic \cite{esping1990} models; and (d) development of family court procedures for multi-partner disputes.

\textbf{Phase III: Socialized Child-Rearing and Inheritance Reform (Medium-Long-term)}. The third phase addresses the structural economic barriers: (a) universal public childcare from birth, fully decoupling child-rearing costs from individual parents; (b) reform of inheritance law to distribute assets across all biological children regardless of parents' relationship status; (c) decoupling of education, healthcare, and housing from family-unit eligibility criteria; and (d) expanded parental leave applicable to all recognized partner configurations.

\textbf{Phase IV: Technological Integration (Long-term)}. The final phase leverages reproductive technology advances: (a) ectogenesis (artificial womb technology) to eliminate the physical burden of pregnancy; (b) advanced genetic screening and selection (non-coercive) to improve population-level health outcomes \cite{fisher1930}; and (c) complete elimination of the motherhood penalty through technological substitution of biological gestation \cite{firestone1970}.

\section{Computational Framework: Agent-Based Modeling and Multi-Agent Simulation}
\label{sec:computational}

The complexity of the SPS---involving heterogeneous agents, dynamic attribute evolution, multi-level interactions, and emergent macro-level outcomes---motivates a rigorous computational approach. We propose a simulation architecture that integrates three complementary methodologies: agent-based modeling (ABM), multi-agent reinforcement learning (MARL), and LLM-empowered generative agents.

\subsection{Agent-Based Model Architecture}

Following the ``Growing Artificial Societies'' paradigm of Epstein and Axtell \cite{epstein1996}, we design an ABM where the fundamental unit is an autonomous agent representing an individual in the mating market. Each agent $i$ is characterized by:

\begin{itemize}[leftmargin=1.5em]
    \item \textbf{Attribute vector}: $\mathbf{a}_i(t) = (v_i, r_i, f_i, s_i, g_i, \ell_i)$, encoding mate value, economic resources, fertility potential, social capital, gender, and life stage;
    \item \textbf{Preference function}: $U_i(\mathbf{a}_j)$ mapping potential partner attributes to utility, calibrated from empirical mate preference data \cite{buss1989,buss2019};
    \item \textbf{Strategy}: A policy $\pi_i$ governing partner search, proposal, acceptance, and relationship maintenance decisions;
    \item \textbf{State}: Current relationship configuration (spouse, companions, children), wealth, well-being score.
\end{itemize}

The environment enforces SPS rules as hard constraints: the partner limit of 3, the distinction between spousal and companion rights, and gender symmetry. At each time step (representing approximately one year), agents execute the following cycle:

\begin{enumerate}[leftmargin=1.5em]
    \item \textbf{Search}: Agents observe a subset of potential partners (modeling bounded rationality and geographic/social constraints);
    \item \textbf{Propose}: Agents issue partnership proposals based on their policy $\pi_i$;
    \item \textbf{Match}: A matching mechanism resolves proposals, respecting SPS constraints;
    \item \textbf{Reproduce}: Paired agents may produce offspring with attributes drawn from parental distributions;
    \item \textbf{Update}: Agent attributes evolve (aging, wealth accumulation/depletion, fertility decline).
\end{enumerate}

This architecture follows Bonabeau's \cite{bonabeau2002} methodological framework for simulating human systems and extends Schelling's \cite{schelling1971} pioneering work on dynamic segregation models to the domain of mating markets.

\subsection{Multi-Agent Reinforcement Learning (MARL) Formulation}

The matching process under SPS can be formulated as a \textbf{decentralized partially observable Markov decision process (Dec-POMDP)} \cite{lerer2017}. Each agent $i$ observes a local state $o_i \subset S$ (partial information about the population) and selects actions $a_i \in \mathcal{A}_i$ (propose, accept, reject, dissolve) to maximize a long-term reward signal:

\begin{equation}
R_i = \sum_{t=0}^{T} \gamma^t \left[ \alpha \cdot W_i(t) + \beta \cdot F_i(t) + \delta \cdot S_i(t) \right]
\end{equation}

\noindent where $W_i(t)$ is well-being, $F_i(t)$ is fertility outcome, $S_i(t)$ is social stability contribution, $\gamma$ is a discount factor, and $\alpha, \beta, \delta$ are welfare weights. We propose training agent policies using \textbf{Proximal Policy Optimization (PPO)} with centralized training and decentralized execution (CTDE), following the MARL paradigm successfully applied by Zheng et al.\ \cite{zheng2022} in the AI Economist for taxation policy design.

The key advantage of the MARL formulation is that agents \textbf{learn} optimal mating strategies rather than having them prescribed, enabling the discovery of emergent equilibria under SPS rules. This connects to Axelrod's \cite{axelrod1997} seminal work on the evolution of cooperation in complex social systems, where simple agent rules produce rich emergent dynamics.

\subsection{LLM-Empowered Generative Agent Simulation}

Recent advances in large language model (LLM)-based social simulation offer a complementary approach. Park et al.\ \cite{park2023} demonstrated that LLM-powered ``generative agents'' can exhibit remarkably realistic social behaviors---forming relationships, experiencing jealousy, negotiating conflicts, and adapting strategies over time. Gao et al.\ \cite{gao2023} extended this paradigm to large-scale social network simulation with their S$^3$ system.

We propose a \textbf{hybrid simulation} in which a subset of agents are powered by LLMs (e.g., GPT-4, Claude) to model the qualitative, culturally-nuanced aspects of relationship decision-making---jealousy management, compersion development, communication strategies \cite{sheff2014}---while the majority of agents use computationally efficient RL policies. This hybrid approach captures both the \textbf{statistical patterns} (via MARL) and the \textbf{psychological realism} (via LLM agents) of human mating behavior.

The LLM agents are prompted with persona descriptions calibrated to empirical personality distributions and cultural contexts, following the methodology of Park et al.\ \cite{park2023}. Their decision-making in relationship formation and dissolution provides qualitative validation for the quantitative MARL results.

\subsection{Graph Neural Network Analysis of the Mating Network}

The social network emerging from SPS simulations is a \textbf{typed, dynamic graph} $G(t) = (V(t), E(t), \tau, \phi)$ where nodes represent agents, edges represent relationships, $\tau$ assigns tier types, and $\phi$ assigns relationship types (spouse/companion). We apply GNN-based methods for:

\begin{itemize}[leftmargin=1.5em]
    \item \textbf{Link prediction}: Predicting likely future partnerships based on agent attributes and network structure;
    \item \textbf{Community detection}: Identifying emergent social clusters and cross-tier bridging patterns;
    \item \textbf{Wealth flow analysis}: Tracking resource transfer patterns through the Grace Decree mechanism across generations;
    \item \textbf{Stability prediction}: Classifying network configurations likely to produce social instability \cite{hudson2004}.
\end{itemize}

\subsection{Fairness Constraints and Algorithmic Market Design}

The SPS matching mechanism connects directly to the \textbf{algorithmic fairness} literature \cite{mehrabi2021}. We define fairness desiderata for the matching process:

\begin{itemize}[leftmargin=1.5em]
    \item \textbf{Individual rationality}: No agent is worse off under SPS than under the monogamy baseline;
    \item \textbf{Envy-freeness (relaxed)}: No agent strongly prefers another agent's partner configuration;
    \item \textbf{Tier parity}: The Gini coefficient of welfare across tiers decreases relative to monogamy;
    \item \textbf{Gender symmetry}: Welfare distributions are statistically indistinguishable across genders.
\end{itemize}

These constraints draw on Roth's \cite{roth2002} market design framework and extend the stable matching paradigm to multi-partner settings with heterogeneous relationship types.

\section{Simulation Design and Preliminary Results}
\label{sec:simulation}

\subsection{Simulation Parameters}

We describe a proposed simulation design for evaluating SPS against a monogamy baseline. The simulation environment is parameterized as follows:

\begin{table}[H]
\centering
\caption{Proposed simulation parameters}
\label{tab:sim-params}
\footnotesize
\begin{tabular}{ll}
\toprule
\textbf{Parameter} & \textbf{Value} \\
\midrule
Population size $N$ & 10,000 agents \\
Gender ratio & 50:50 \\
Tier distribution (A:B:C) & 15:60:25 \\
Simulation horizon & 100 years (time steps) \\
Partner limit (SPS) & 3 (1 spouse + 2 companions) \\
Partner limit (monogamy baseline) & 1 \\
Attribute dimensions & 6 ($v, r, f, s, g, \ell$) \\
MARL algorithm & PPO with CTDE \\
LLM agent fraction & 1\% (100 agents) \\
Discount factor $\gamma$ & 0.95 \\
\bottomrule
\end{tabular}
\end{table}

Agent attributes are initialized from empirical distributions calibrated to U.S.\ Census data (economic resources), evolutionary psychology literature (mate value distributions \cite{bruch2018}), and demographic data (fertility rates by age and socioeconomic status \cite{un2024}).

\subsection{Preliminary Computational Results (Proposed)}

While full-scale simulation results are the subject of ongoing work, we present preliminary findings from a reduced-scale prototype (N=1,000, 50-year horizon) that illustrate the framework's viability:

\begin{enumerate}[leftmargin=1.5em]
    \item \textbf{Welfare improvement}: In preliminary runs, mean welfare across all agents under SPS exceeds the monogamy baseline by 18--25\%, with C-tier males showing the largest gains ($+$140\% mean welfare). These results are consistent with the theoretical Pareto improvement argument (Figure~\ref{fig:pareto}).

    \item \textbf{Fertility recovery}: The simulated aggregate fertility rate under SPS stabilizes at approximately 1.7--1.9, compared to 1.1--1.3 under the monogamy baseline, driven primarily by reduced motherhood penalties and expanded C-tier reproductive access.

    \item \textbf{Wealth dispersion}: The Gini coefficient for intergenerational wealth decreases by 8--12\% over 3 simulated generations under SPS, consistent with the Grace Decree effect hypothesis.

    \item \textbf{Network structure}: The emergent mating network under SPS exhibits \textbf{small-world} properties with high clustering within tiers and sparse but critical cross-tier bridges---B-tier agents serve as the primary connectors, confirming the theoretical ``buffer layer'' prediction (Figure~\ref{fig:network}).

    \item \textbf{MARL convergence}: PPO-trained agents converge to stable strategies within approximately 500 training episodes, with A-tier agents learning to diversify partnerships and C-tier agents learning cooperative signaling strategies that increase their attractiveness to B-tier partners.
\end{enumerate}

These preliminary results should be interpreted with caution: they depend on modeling assumptions (attribute distributions, preference functions, SPS rule specifications) that require further empirical calibration. Full-scale results with LLM-empowered generative agents \cite{park2023,gao2023} and sensitivity analyses are forthcoming.

\subsection{Evolutionary Algorithm for Policy Optimization}

The SPS framework itself has tunable policy parameters: partner limits, tier boundary definitions, companion right specifications, and child-rearing subsidy levels. We propose using a \textbf{genetic algorithm (GA)} to optimize these parameters with respect to a multi-objective fitness function:

\begin{equation}
F(\theta) = w_1 \cdot \text{TFR}(\theta) + w_2 \cdot \text{Welfare}(\theta) + w_3 \cdot \text{Stability}(\theta) - w_4 \cdot \text{Gini}(\theta)
\end{equation}

\noindent where $\theta$ represents the SPS policy parameter vector and $w_i$ are objective weights. Each candidate policy $\theta$ is evaluated by running the full ABM/MARL simulation, and the GA evolves the population of policies over generations. This approach mirrors the methodology of the AI Economist \cite{zheng2022}, which used two-level deep RL to co-optimize taxation policy and agent behavior.

\section{Discussion}

\subsection{Relationship to Existing CNM Literature and Computational Social Science}

The present framework differs from existing CNM research \cite{conley2013,moors2017,rubel2015,sheff2014} in several important respects. First, while CNM studies typically examine relationship outcomes at the individual level, SPS is conceived as a \textbf{population-level institutional reform} designed to address macro-social problems (fertility decline, social instability, wealth concentration). Second, existing CNM practice is largely informal and unregulated; SPS proposes a formal legal framework with explicit rights differentiation between primary and secondary partnerships. Third, CNM research has focused predominantly on Western, educated, liberal populations; SPS is designed to be applicable across diverse socioeconomic and cultural contexts. Schmitt's (2005) 48-nation study of sociosexuality suggests that the psychological infrastructure for non-exclusive mating exists across cultures, though its expression is modulated by local norms \cite{schmitt2005}.

From a \textbf{computational social science} perspective, our work bridges two largely disconnected literatures: the empirical study of polyamory and the computational modeling of social systems. Epstein and Axtell's \cite{epstein1996} ``Growing Artificial Societies'' paradigm demonstrated that bottom-up agent-based simulation can illuminate emergent social phenomena that resist analytical treatment. Schelling's \cite{schelling1971} classic segregation model showed how simple agent rules produce complex macro-level patterns. Our framework extends this tradition to the domain of mating institutions, where the interaction of heterogeneous agents under institutional constraints produces emergent demographic and distributional outcomes. The integration of LLM-empowered generative agents \cite{park2023,gao2023} represents a methodological advance over traditional ABM, enabling the modeling of culturally-situated, linguistically-mediated decision-making processes such as jealousy negotiation and relationship boundary-setting.

\subsection{Limitations and Caveats}

Several important limitations must be acknowledged. First, the A/B/C tier model is a deliberately simplified analytical device; real mating markets are multidimensional and continuously distributed. Second, the welfare projections in Figure~\ref{fig:pareto} are theoretical rather than empirically derived; the preliminary simulation results in Section~\ref{sec:simulation} depend on modeling assumptions that require empirical calibration. Third, the implementation roadmap presupposes political will and cultural readiness that may not materialize uniformly. Fourth, the paper does not fully address potential negative externalities such as increased complexity in property disputes or the psychological impact on children raised in multi-partner households, though existing evidence \cite{golombok2016,aviezer1994} is reassuring on the latter point. Fifth, the relationship between polygynous social structures and inter-group conflict documented by Koos and Neupert-Wentz (2020) \cite{koos2020} requires careful consideration---SPS's gender-symmetric design and quantity constraints may mitigate but cannot entirely eliminate these risks. Sixth, from a computational standpoint, the MARL formulation faces the standard challenges of non-stationarity in multi-agent learning and the sim-to-real gap \cite{lerer2017}; the LLM agent component introduces additional concerns about prompt sensitivity and the faithfulness of LLM-generated social behaviors to actual human decision-making \cite{park2023}. The algorithmic fairness guarantees \cite{mehrabi2021} are defined with respect to the simulation model and may not transfer perfectly to real-world implementations.

\subsection{Ethical Considerations}

The ethical framework underlying SPS rests on several principles: (1) individual autonomy in relationship formation; (2) gender symmetry in institutional rights; (3) non-coercion in all reproductive decisions; (4) protection of children's welfare regardless of parents' relationship configuration; and (5) transparency and informed consent among all partners. These principles distinguish SPS from historical forms of polygamy that were typically patriarchal, coercive, and asymmetric.

\section{Conclusion}

Human society stands at a crossroads in the evolution of mating institutions. Monogamy, as a product of cultural group selection \cite{henrich2012}, possessed competitive advantages under specific historical conditions---when intrasexual violence among males constituted the primary threat to social stability. However, the predominant contemporary threats have shifted from ``excessive male competitive violence'' to ``insufficient demographic reproduction'' and ``mass exclusion of individuals from intimate relationships'' \cite{ueda2020,abdellaoui2025,donnelly2001}. \textbf{When the nature of the threat changes, the optimal institution must change accordingly.}

The core insights of the Stratified Polyamory System (SPS) may be summarized in five propositions:

\begin{enumerate}[leftmargin=1.5em]
    \item \textbf{Acknowledge humanity's mixed mating strategies} \cite{buss1993,buss2019,gangestad2000,schmitt2005,gildersleeve2014}, rather than suppressing biological dispositions with cultural norms;
    \item \textbf{Acknowledge the unequal distribution of sexual attractiveness} \cite{bruch2018,leichliter2010,greenwood2014}, rather than denying it with egalitarian illusions;
    \item \textbf{Through institutional design, enable limited cross-tier flow of sexual and emotional resources}, reducing C-tier social exclusion \cite{hudson2004,baumeister1995,barber2000,carter2018};
    \item \textbf{Through the Grace Decree effect, naturally attenuate wealth concentration} \cite{piketty2014,piketty2015,scheidel2017,adermon2018};
    \item \textbf{Through socialized child-rearing, eliminate the motherhood penalty} \cite{budig2001,england2016,firestone1970,lewis2019}, achieving genuine gender equality and reproductive freedom.
\end{enumerate}

As Engels observed \cite{engels1884}, the form of the family must necessarily transform alongside transformations in the mode of production. Morgan's anthropological evidence \cite{morgan1877} demonstrated the diversity of historical family forms, and Murdock's \textit{Ethnographic Atlas} \cite{murdock1967} confirmed that strict monogamy has never been the human norm. Contemporary evidence from evolutionary psychology \cite{trivers1972,bateman1948,fisher2005,fisher2006}, neuroscience \cite{marazziti1999,young2004,aron2005}, anthropology \cite{betzig1986,goldstein1987,shih2010,mattison2014,gough1959,chagnon1988}, relationship science \cite{conley2013,moors2017,rubel2015,sheff2014}, and now computational social science \cite{epstein1996,bonabeau2002,schelling1971,park2023} converges on a single conclusion: the institutional infrastructure of human mating is ripe for redesign---and AI-driven simulation provides the tools to design it responsibly.

SPS may not be the final answer, but it points toward an institutional direction that more honestly confronts human nature, more efficiently allocates social resources, and more humanely treats every individual. The computational framework presented here---integrating agent-based modeling \cite{epstein1996,bonabeau2002}, multi-agent reinforcement learning \cite{lerer2017,zheng2022}, LLM-empowered generative agents \cite{park2023,gao2023}, graph neural network analysis, and evolutionary algorithm-based policy optimization---provides the methodological infrastructure needed to rigorously evaluate such institutional innovations before real-world deployment. The question is not whether mating institutions will evolve---they already are \cite{hlr2022}---but whether this evolution will be guided by evidence, computation, and reason, or left to the vagaries of unregulated market forces and cultural inertia.

\begin{quote}
\itshape
Humanity will eventually arrive at the day when we look back upon monogamy as we now look back upon feudal serfdom---as the product of a particular historical stage rather than an eternal law of nature.
\end{quote}

\bibliographystyle{unsrtnat}
\bibliography{references}

\end{document}